\pdfoutput=1

\documentclass[11pt]{article}

\usepackage[preprint]{ACL2025/acl}
\usepackage{times}
\usepackage{latexsym}

\usepackage{booktabs}
\usepackage{multirow}
\usepackage{amssymb}
\usepackage{amsmath}
\usepackage{pifont}
\usepackage{wrapfig}
\usepackage{subcaption}

\usepackage{enumitem}

\usepackage[T1]{fontenc}

\usepackage[utf8]{inputenc}

\usepackage{microtype}

\usepackage{inconsolata}

\usepackage{graphicx}
\usepackage{comment}
\usepackage{xspace}

\newcommand{\jo}[1]{{\color{black}{#1}}}
\newcommand{\geon}[1]{{\color{black}{#1}}}

\newcommand{\ours}{\textit{Talking with Tables}\xspace}
\newcommand{\prompt}[1]{{\small \ttfamily #1}\xspace}

\usepackage{todonotes}
\usepackage{xcolor}

\title{Talking with Tables for Better LLM Factual Data Interactions}

\author{
  \textbf{Jio Oh\footnotemark[1]\footnotemark[3]\textsuperscript{1}\textsuperscript{2}},
  \textbf{Geon Heo\footnotemark[1]\textsuperscript{1}},
  \textbf{Seungjun Oh\textsuperscript{1}},
  \textbf{Hyunjin Kim\textsuperscript{2}\textsuperscript{4}},\\
  \textbf{JinYeong Bak\textsuperscript{4}},
  \textbf{Jindong Wang\textsuperscript{3}},
  \textbf{Xing Xie\textsuperscript{2}},
  \textbf{Steven Euijong Whang\footnotemark[2]\textsuperscript{1}}\\\\
\textsuperscript{1}KAIST,
  \textsuperscript{2}Microsoft Research Asia
  \textsuperscript{3}William \& Mary
  \textsuperscript{4}SKKU
  }

\begin{document}

\maketitle


\begin{abstract}
  Large Language Models (LLMs) often struggle with requests related to information retrieval and data manipulation that frequently arise in real-world scenarios under multiple conditions. In this paper, we demonstrate that leveraging tabular structures in LLM interactions, is more effective than utilizing other structures for handling prevalent requests that operate over factual data. Through comprehensive evaluations across various scenarios and request types, we show that providing tabular structures yields a 40.29\% average performance gain along with better robustness and token efficiency. Through attention-value analysis, we discover that tables help LLMs better locate relevant information, explaining these improvements. Beyond tables and text, we evaluate whether (1) blending structuredness within text, such as providing templates or fixing the order of attributes, and (2) other representative structures, such as knowledge graphs and JSON are helpful. We observe that utilizing tables offers the best balance between efficiency and effectiveness.  The method remains robust to task complexity and adapts to unstructured sources through text-to-table conversion. Overall, we highlight the untapped potential of tabular representations for future LLM applications.

\end{abstract}

\section{Introduction}
\label{sec: intro}

\begin{figure}[t]
    \centering
    \includegraphics[width=0.97\linewidth]{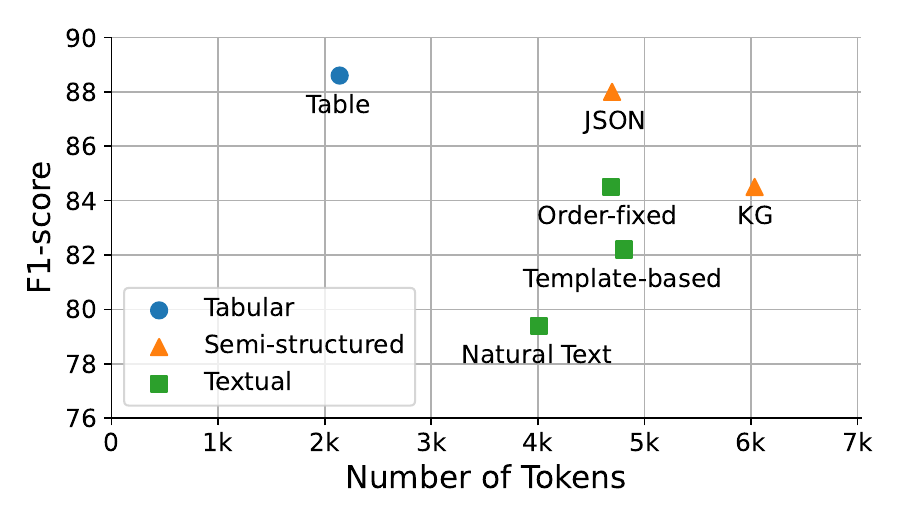}
    \captionsetup{skip=4pt}
    \caption{\geon{Comparison of different structures in terms of LLM performance and the number of tokens. Although other formats include additional information about the context or relationship between attributes, the {\em Table} format achieves the highest performance, indicating token efficiency and effectiveness. Textual formats are divided into three different structuring levels (Green). The examples of each structure are in Tbl.~\ref{tab: structuring level} and \ref{tab: semi-structure_representation}, and the extensive results are in Sec.~\ref{sec: result}.}} 
    \vspace{-0.1cm}
    \label{fig:motivation}
\end{figure}

The recent advancement of Large Language Models (LLMs) has transformed the field of natural language processing, where LLMs are serving as alternatives to traditional search engines \citep{wang-etal-2024-factuality}. With studies showing that 50-60\% of web queries are focused on informational queries, related to retrieving factual data under specified constraints \citep{10.1145/792550.792552, 10.1145/988672.988675, 10.1145/1753846.1754140}, it is increasingly important for LLMs to show high performance on these types of requests. 

\begin{figure*}[ht]

    \centering
    \includegraphics[width=\linewidth]{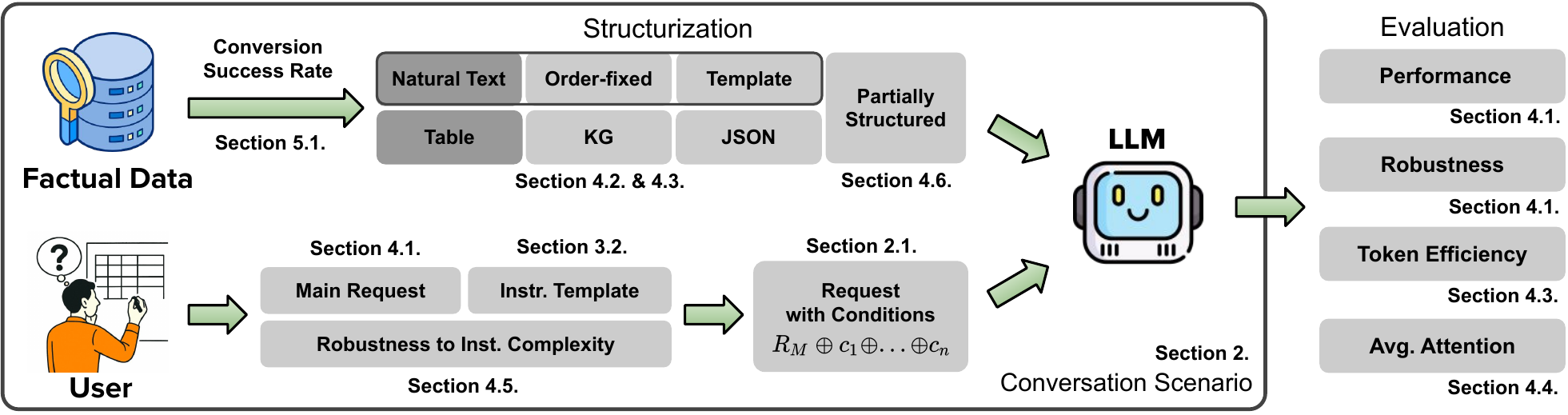}
    \captionsetup{skip=4pt}
    \caption{Evaluation Framework. We design to evaluate the impact of structures on LLMs’ performance and robustness for requests that operate over factual data, with analyses and ablation studies across various models, examining the generalizability of \ours{}. We compare LLM performances with different structural representation of the same information in scenarios shown in Fig.~\ref{fig:scenario}.} 
    \vspace{-0.2cm}
    \label{fig:framework}
\end{figure*}
For instance, users commonly ask questions like ``What was the movie directed by Christopher Nolan and starring Cilian Murphy?'' or ``How many American players are in the English Premier League?''. These requests, often including information retrieval, aggregation,  or data manipulation on factual data~\citep{barke2024solvingdatacentrictasksusing}, require models to correctly identify entities satisfying certain conditions~\citep{sen-etal-2022-mintaka}, a task that current state-of-the-art LLMs often struggle with, achieving F1 scores below 70\% in our experiments (Tbl.~\ref{tbl:promptperformance}).

Moreover, several studies have shown that LLMs tend to struggle with complex requests involving additional details, leading to a noticeable decline in performance as the query complexity increases~\citep{he2024can,xiong2020answering,zhang2024cfbench}. 
Such requests are seldom found in text used to train LLMs, making it even more challenging for the model to generate accurate responses~\citep{he2024complex,zhou2023lima}. 

Building on the challenges above, it becomes increasingly important to equip LLMs with the capability to handle these requests effectively. One of the major factors of wide usage of LLMs is its capability of multi-turn interactions. However, LLMs are highly susceptible to snowball effects, which causes hallucinations.~\cite{zhang2023language,oh2024erbench}. An initial error propagates through conversations, causing the model to deviate from the correct state. Hence, ensuring high precision in initial factual tasks, such as retrieval, is critical, as early-stage performance determine the reliability of the entire subsequent interactions.

When providing factual data to LLMs, most studies typically rely on natural language text~\citep{lewis2020retrieval} or semi-structured formats such as JSON objects~\citep{schick2023toolformer} and knowledge graphs~\cite{markowitz2025kg}. This has become the standard in both research and deployed systems, with considerable effort devoted to information extraction around these formats. However, a fundamental question remains underexplored: does the structural format in which data is presented to LLMs affect their performance? Specifically, tables, being the most structured format for factual data with their two-dimensional layout and high token efficiency, have received surprisingly little attention as an input representation.



Tables contain information structured into a two-dimensional grid with fixed column names and positions,  highlighting relationships and conditions at a glance. In contrast, one-dimensional textual representations, which are often accompanied with diverse expressions and phrasing, can introduce inconsistencies and inaccuracies in LLM responses~\cite{sclar2023quantifying,errica2024did,bonagiri2024sage}. This structural advantage aligns with human cognitive behaviors, which utilizes tabular organization to identify patterns and relationships between variables~\citep{blanton2011functional, cloutier2021using}. This suggests that LLMs similarly leverage such organization to untangle complexity. Moreover, tables also present information in a compact and succinct manner, significantly reducing the number of tokens needed to convey equivalent information compared to text, JSON, or knowledge graphs. As a result, tables allow LLMs to ingest more factual information per context window and lower inference cost. This dual advantage of effectiveness and efficiency makes tables particularly attractive given the finite context windows and inference costs of modern LLMs.

In this paper, we conduct a comprehensive empirical study comparing LLM performance across different data representation formats: natural text, text with varying levels of structure (order-fixed, template-based), JSON objects, knowledge graphs, and tables. Specifically, we investigate requests that operate over factual data, where tabular formats are prevalent as a primary medium for representing and storing structured information. We find that tabular structures consistently improve LLM performance, achieving a 40.29\% average relative improvement over natural text—while also being the most token-efficient format (Fig.~\ref{fig:motivation}).



We further provide analyses to characterize when and why tables help. First, tables reduce output variance across semantically equivalent prompts, indicating improved robustness. Second, attention analysis reveals that tables help models better focus on relevant information, offering a mechanistic explanation for the performance gains. Third, the benefits persist across varying numbers of conditions, different context lengths, and even when only a portion of the data can be structured. Notably, tables excel in sparse data scenarios where attributes are partially missing, a setting where JSON and knowledge graphs actually degrade performance below that of plain text.

\begin{figure*}[ht]
    \centering
    \includegraphics[width=\linewidth]{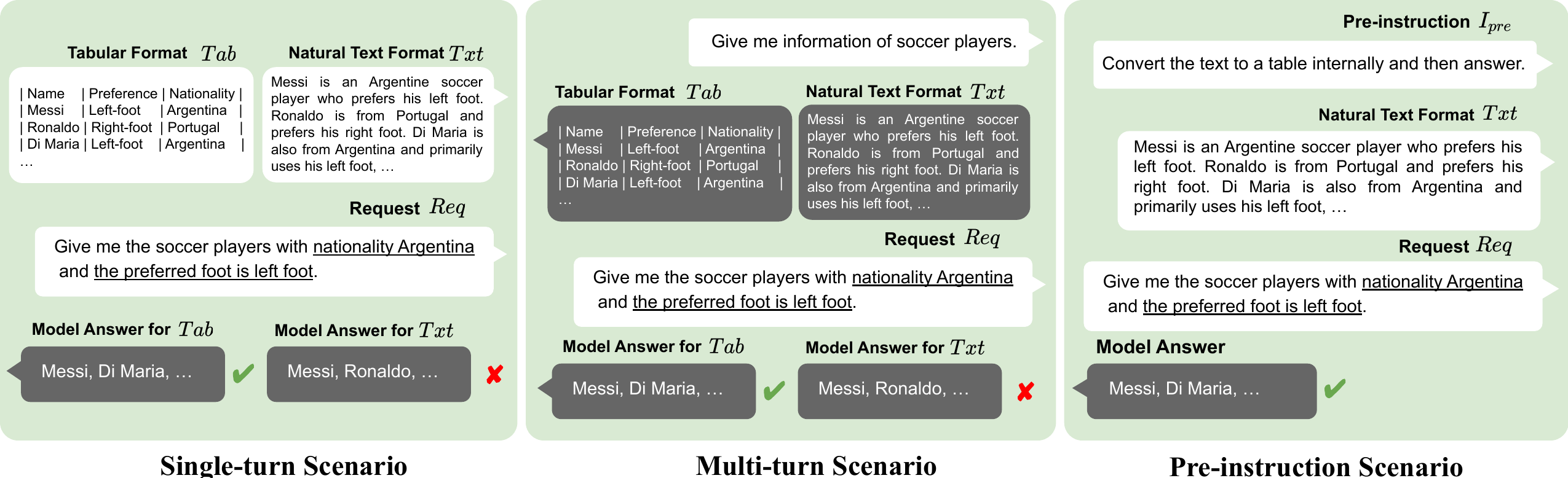}
    \captionsetup{skip=4pt}
    \caption{To evaluate the effectiveness of structuredness, we define and categorize three types of conversation scenarios: \textit{single-turn}, \textit{multi-turn}, and \textit{pre-instruction}. A request $Req$ consists of a main request and additional conditions (underline). White and black speech bubbles denote user requests and model responses, respectively.} 
    \label{fig:scenario}
\end{figure*}

Our contributions are as follows: (1) We show empirically that tabular structures improve LLM accuracy and robustness while being the most token-efficient representation compared to other structures. (2) We provide extensive analyses across multiple models, scenarios, and ablation studies that characterize the conditions under which tabular representations offer advantages. (3) We highlight the untapped potential of overlooked tabular representations, providing guidance for future applications. In Fig.~\ref{fig:framework}, we summarize our whole evaluation framework.

\section{Evaluation Framework} 
\label{sec: method}

Despite the high-quality responses of LLMs for general requests, they often output incorrect results on requests on factual data with detailed constraints~\citep{he2024complex,he2024can,barke2024solvingdatacentrictasksusing}. These requests often include information retrieval and data manipulation such as filtering, aggregation, and transformation that are prevalent in real-world user interactions and widely present in various benchmarks~\cite{joshi-etal-2017-triviaqa,sen-etal-2022-mintaka, srivastava2023beyond, li-etal-2023-halueval, oh2024erbench,chen2024factchd, sun-etal-2024-head}. Ensuring correctness for these requests is crucial as LLM-based data-analytics allows higher accessibility to non-experts (see Appendix~\ref{app:motivation} for details). In this paper, we show structuredness, representatively tables, boost models' answer accuracy and robustness, while being the most efficient. 



\subsection{Request Formalization.}

To scale the diversity of our evaluation set, we decompose each request into two components: a main request and conditions. For instance, an instruction can be constructed by concatenating a main request (``\prompt{Give me the number of soccer players}'') with specific conditions (``\prompt{who play in the English Premier League}'' and ``\prompt{born in USA}''). Generally, when a request $Req$ is inputted to an LLM, we express the model output $A_{0}$ as follows:
{\setlength{\abovedisplayskip}{4.5pt}
 \setlength{\belowdisplayskip}{4.5pt}
\begin{equation*}
\label{eq:rdc}
    A_{0} = LLM(Req)\;\;\;Req:=R_M \oplus c_1 \oplus ... \oplus c_n
\end{equation*}}
where $R_M$ denotes the main request, and $c_{i\in[1...n]}$ denotes $i$th condition among the $n$ additional conditions. $\oplus$ represents the concatenation~\citep{he2024complex}. For a request type with $n$ conditions and a pool of $m$ candidates (where $m \gg n$), the number of possible instructions scales in a  combinatorial manner as $n \cdot \binom{m}{n}$. This space is further expanded by varying main request types and utilizing different logical compositions, such as ``and'' and ``or'' operators. Details about generating conditions are provided in Appendix~\ref{app: cond}.

\begin{table*}[th]
    \centering
    \caption{Data structuring levels. To evaluate the influence of data structuredness, we further subdivide the levels by incrementally adding structuredness to text. In this paper, “Text” refers to \textit{Natural} text. }
    \scalebox{0.75}{
    \begin{tabular}{c|c|c|c|l}
    \toprule
     & Fixed & Fixed & Tabular & \multirow{2}{*}{Example}\\
     & Order & Expression & Format & \\
    \midrule
    \midrule
    Natural & \multirow{2}{*}{\ding{55}} & \multirow{2}{*}{\ding{55}} & \multirow{2}{*}{\ding{55}} & \textbf{Ronaldo}, a player for \textbf{Juventus}, wore \textbf{jersey number 7} and is from \textbf{Portugal}.\\
    (Text)& & & & \textbf{Messi} is a player from \textbf{Argentina} playing for \textbf{Barcelona} with \textbf{uniform number 10}.\\
    \midrule
    \multirow{2}{*}{Order-fixed} & \multirow{2}{*}{\checkmark} & \multirow{2}{*}{\ding{55}} & \multirow{2}{*}{\ding{55}} &\textbf{Ronaldo}, a \textbf{Portuguese} player, was with \textbf{Juventus}, while wearing \textbf{jersey No. 7}.\\
    & & & & \textbf{Messi} is a player from \textbf{Argentina} playing for \textbf{Barcelona} with \textbf{uniform number 10}.\\
    \midrule
    \multirow{2}{*}{Template-based} & \multirow{2}{*}{\checkmark} & \multirow{2}{*}{\checkmark} & \multirow{2}{*}{\ding{55}} &\textbf{Ronaldo} is a player from \textbf{Portugal} playing for \textbf{Juventus} with \textbf{uniform number 7}. \\
    & & & & \textbf{Messi} is a player from \textbf{Argentina} playing for \textbf{Barcelona} with \textbf{uniform number 10}.\\
    \midrule
    \multirow{3}{*}{Table} & \multirow{3}{*}{\checkmark} & \multirow{3}{*}{\checkmark} & \multirow{3}{*}{\checkmark} & | Name | Number | Nationality | Club |\\
    & & & & | \textbf{Ronaldo} | \textbf{7} | \textbf{Portugal} | \textbf{Juventus} |     \\
    & & & & | \textbf{Messi} | \textbf{10} | \textbf{Argentina} | \textbf{Barcelona} |     \\
    \bottomrule
    \end{tabular}
    }
    \label{tab: structuring level}
\end{table*}

\subsection{Scenario Formalization.} 
\label{subsec:scenario_formalization}
To reflect realistic usage patterns, we consider different scenarios of how users interact with and retrieve factual data. For instance, the associated factual data can be provided by the user, whereas the user might require the model to first retrieve the relevant data. To comprehensively evaluate the effectiveness of structuredness, we define three types of scenarios: \textit{single-turn}, \textit{multi-turn}, and \textit{pre-instruction}. Fig.~\ref{fig:scenario} shows simplified examples for each scenario. 



\noindent\textbf{ 1) Single-turn:}
Comparison of cases where table and text form of data is \textit{inputted to} the model.
{\setlength{\abovedisplayskip}{4.5pt}
 \setlength{\belowdisplayskip}{4.5pt}
\begin{equation*}
    LLM(Tab,\; Req ) \Leftrightarrow LLM(Txt, \; Req)
    \label{eq: single_1}
\end{equation*}}
\noindent\textbf{2) Multi-turn:}
Comparison of cases where table and text form of data is \textit{outputted by} the model.
{\setlength{\abovedisplayskip}{4.5pt}
 \setlength{\belowdisplayskip}{4.5pt}
\begin{equation*}
    LLM(Req | \;Tab) \Leftrightarrow LLM(Req | \;Txt)
\end{equation*}}
\noindent\textbf{3) Pre-instruction:}
Comparison of cases where an \textit{additional instruction} to structurize information is given to the model with text form of data.
{\setlength{\abovedisplayskip}{4.5pt}
 \setlength{\belowdisplayskip}{4.5pt}
\begin{equation*}
    LLM(Txt, \; I_{pre}, \; Req) \Leftrightarrow LLM(Txt, \; Req )
    \label{eq: single_2}
\end{equation*}}
where $Tab$ and $Txt$ denote data in tabular and textual formats, respectively, $Req$ denotes a user request, and $I_{pre}$ is a simple prefix instruction as \prompt{``Convert the text to a table internally and then answer.''}. We set the \textit{single-turn} scenario as our main experiments to eliminate unexpected prompt-engineering effects, which might occur due to prompts like \prompt{``Give me information of soccer players''} in Fig.~\ref{fig:scenario}. This scenario allows us to directly isolate the impact of data structures.

\subsection{Extended Data Formats.} Beyond conventional textual and tabular formats, we construct intermediate data structuring levels by interpolating between natural text and table. We add structuredness to text by generating these representations using fixed templates and paraphrasing sentences while maintaining the order of attributes (see Sec.~\ref{subsec:data structuring} for details). We further extend our analysis to other popular structured formats: knowledge graphs (KGs) and JSON (see Sec.~\ref{subsec:why_tables} and Appendix~\ref{app:representation_semi_struct} for details). Across all data formats, we ensure information equivalence by removing surplus attributes, ensuring that each representation contains the exact same set of attributes (Appendix ~\ref{app: fair comparison}). Tbl.~\ref{tab: structuring level} provides illustrative examples of different data structuring levels.

\section{Experimental Setup} \label{sec: setup}

In this section, we provide the models and datasets utilized in our experiments. We describe the specific types of requests formulated to assess the model's performance and outline the evaluation criteria applied to each request. Following \citet{jin2023tab}, we use the ``|'' delimiter for table formatting, as it has been shown to outperform alternatives like commas in LLM-based interpretation.

\subsection{Models and Datasets}
\label{subsec:llmsdatasets}
To thoroughly evaluate our approach, we assessed the top-performing models, five proprietary (GPT-3.5, GPT-4~\citep{openai2024gpt4technicalreport}, GPT-4o, Gemini-1.5-Flash~\citep{gemini}, and Claude-3.5-Sonnet) and three open-source models (LLaMA-3.1-70B, Mixtral-8x22B. Gemma-2-27B). For the \textit{pre-instruction} scenario, we test two reasoning models (GPT-o3-mini and Gemini-2.5-Flash). We use three datasets presenting different domains: \texttt{Soccer}~\citep{fifa20}, \texttt{Movie}~\citep{imdb}, and \texttt{PII}~\citep{pii} (see Appendix~\ref{app: dataset} for details). \jo{We standardize the input data by fixing the number of entities to match the smallest maximum context window among the evaluated LLMs.} 

\subsection{Request Types and Evaluation Metrics}
\label{subsec: task}

In this section, we show how we select main requests and evaluation metrics in the experiments. 


\noindent\textbf{Request Types.}
We evaluate six representative request types on factual data, inspired by complex Q\&A and complex instruction benchmarks~\citep{sen-etal-2022-mintaka,he2024can,zhang2024cfbench}. These request types cover core operations over factual data, present in other popular benchmarks as mentioned in Sec.~\ref{subsec:scenario_formalization}. Moreover, the request types can be expanded with database querying languages (see Sec.~\ref{subsec:ext} for details). Examples for each request type is shown in Tbl.~\ref{tbl:promptrelationalalgebra}.



\begin{itemize}[leftmargin=1.2em]
    \vspace{-0.25cm}
    \item\textbf{Retrieval:} Requests to retrieve entities satisfying given conditions.
    \vspace{-0.25cm}
    \item\textbf{Exclusion:} Requests to delete all information about entities satisfying given conditions.
    \vspace{-0.25cm}
    \item\textbf{Revision:} Requests to redact specific information of entities satisfying given conditions.
    \vspace{-0.25cm}
    \item\textbf{Superlative:} Requests to retrieve information based on an entity’s position in an ordered list.
    \vspace{-0.25cm}
    \item\textbf{Summation:} Requests to calculate the sum of the specific information (numerical) for entities satisfying given conditions.
    \vspace{-0.25cm}
    \item\textbf{Quantification:} Requests to count the number of entities satisfying given conditions.
    \vspace{-0.25cm}
\end{itemize}
For each request type, we generate three semantically equivalent prompts (e.g., \prompt{``Give/Provide/Show me information''} for the \textit{Retrieval} request)  for comprehensive evaluation. This approach mitigates the risk of performance bias on a certain template, enhancing the robustness of our evaluation. We generate 100 condition pairs each concatenated with both ``and'' and ``or'' for all templates, resulting in a total of 600 instructions for each request type.


\noindent\textbf{Evaluation Metrics.} 
We employ different metrics tailored to each request type to fairly evaluate model performance. For tasks where true/false positives and negatives are applicable (\textit{Retrieval}, \textit{Exclusion}, and \textit{Revision}), we calculate F1 scores. We compute accuracy measures for requests that require a single answer (\textit{Superlative} and \textit{Summation}), whereas for \textit{Quantification}, we compute absolute difference values to account for cases where the model's output is close to the gold answer. This approach allows better performance evaluation over the traditional answer-based accuracy metrics.

\section{Results} \label{sec: result}

\begin{table*}[th]
    \caption{Performance comparison in the \textit{single-turn} scenario across three datasets. Tabular structure results in better performance in most cases. (“Text”, being the baseline,  refers to \textit{Natural} text)}
    \vspace{-0.1cm}
    \label{tbl:promptperformance}
    \centering
    \scalebox{0.9}{
    \begin{tabular}{c|c|c@{\hspace{8pt}}c@{\hspace{8pt}}c@{\hspace{8pt}} c@{\hspace{8pt}}c@{\hspace{8pt}}c@{\hspace{8pt}} c@{\hspace{8pt}}c|c}
    \toprule
     Request Type & Data Type & GPT-3.5 & GPT-4 & GPT-4o & Gemini & Claude & Mixtral & Llama & Gemma & Avg. \\
    \midrule
    \midrule
     \bf Retrieval & Text & 19.9 & 67.5 & 65.9 & 54.1 & 72.9 & 49.5 & 59.0 & 55.5 & 55.5 \\ 
     (F1 score, \%) & Table & \textbf{30.0} & \textbf{77.2} & \textbf{71.5} & \textbf{65.1} & \textbf{76.2} & \textbf{54.0} & \textbf{63.5} & \textbf{59.5} & \bf 62.1 \\
     \midrule
     \bf Exclusion & Text & 27.4 & 53.9 & 48.5 & 25.9 & 32.5 & 19.9 & 35.3 & 23.0 & 33.3\\ 
     (F1 score, \%) & Table & \textbf{32.9} & \textbf{55.8} & \textbf{55.8} & \textbf{35.0} & \textbf{34.7} & \textbf{30.3} & \textbf{37.3} & \textbf{30.3} & \bf 39.0\\
      \midrule
      \bf Revision & Text & 2.8 & 14.8 & 16.1 & 21.9 & 2.0 & 3.0 & \bf 5.4  & 2.3 & 8.5 \\ 
      (F1 score, \%) & Table & \bf 10.8 & \bf 21.7 & \bf 32.6 & \bf 70.4 & \bf 12.8 & \bf 3.4 & 5.1 & \bf 7.2 & \bf 20.5\\
      \midrule
      \midrule
     \bf Superlative & Text & 8.2 & 27.2 & 23.3 & \textbf{21.6} & 60.4 & 22.9 & \textbf{22.7} & 12.5 & 24.8\\ 
     (Accuracy, \%) & Table & \bf 9.8 & \textbf{30.9} & \textbf{25.8} & 18.8 & \textbf{67.2} & \textbf{24.2} & 18.4 & \textbf{13.4} & \bf 26.0\\
     \midrule
     \bf Summation & Text & \bf 3.2 & 15.2 & 9.3 & 4.3 & 36.6 & 17.0 & 7.7 & \bf 3.3 & 12.1\\ 
     (Accuracy, \%) & Table & 3.1 & \textbf{16.5} & \textbf{11.4} & \textbf{5.4} & \textbf{39.9} & \textbf{19.1} & \textbf{8.4} & 2.6 & \bf 13.3\\
     \midrule
     \midrule
     \bf Quantification & Text & 8.75 & 5.01 & 3.87 & 6.22 & 3.61 & 7.86 & 7.68 & 8.39 & 6.42\\ 
     (Abs. Diff. $\downarrow$) & Table & \textbf{8.59} & \textbf{3.23} & \textbf{3.52} & \textbf{5.77} & \textbf{2.58} & \textbf{6.29} & \textbf{7.39} & \textbf{6.78} & \bf 5.52\\
    \bottomrule
    \end{tabular}
    \vspace{-0.1cm}
        }

  \end{table*}

\subsection{Main Results}
\label{subsec: main}

We evaluate the impact of tabular formatting across three scenarios as described in Sec.~\ref{subsec:scenario_formalization}: \textit{single-turn} (Tbl.~\ref{tbl:promptperformance}), \textit{multi-turn} (Tbl.~\ref{tbl:multiturnperformance}), and \textit{pre-instruction} (Tbl.~\ref{tbl:preinstruction_res}) and explore how providing structuredness with respect to tables helps model performances. In the \textit{single-turn} scenario, tabular representations outperforms the textual baseline for all request types, with an average relative increase of 40.29\% (5.34 pp, percentage points). Notably, performance on the \textit{Revision} task, being the most complex due to its joint requirements for retrieval and manipulation, increases by 11.96 pp. Plus, GPT-family models which are known to be pretrained with tables~\cite{jin2023tab} benefit significantly. 

Similar trends hold in both \textit{multi-turn} and \textit{pre-instruction} scenarios. Qualitative analysis (Appendix~\ref{subsec:casestudy_multiturn}) demonstrates a case where that tabular formats enable correct outputs where text fails. Models often explicitly identify the need for structured data for efficient processing, reinforcing the value of tables for high-precision factual tasks. For the \textit{pre-instruction} scenario, we prompt the model to convert the given text to a table before answering. To rigorously evaluate these benefits, we assess the performance of reasoning models, LLMs trained to use test-time compute to enhance inference quality~\cite{xu2025towards}. Interestingly, a simple additional prompt, which guides the models to utilize structures, results in performance increase. Detailed results are provided in Appendix~\ref{app:more_main_res} along with ablation studies of evaluating \textit{single-turn} scenario performance across smaller models (1B to 8B) and the benefits of \textit{pre-instruction} for non-reasoning models.

\begin{figure}[tbp]
    \centering
    \includegraphics[width=0.9\linewidth]{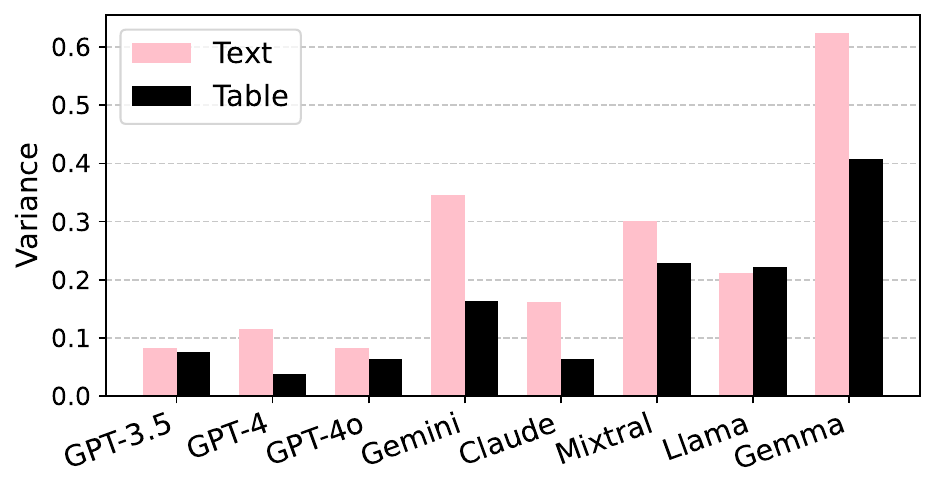}
    \vspace{-0.2cm}
    \caption{Variance of each model's performance across different instruction templates. Using a tabular structure leads to more consistent performance.}
    \vspace{-0.2cm}
    \label{fig:var}
\end{figure}

Beyond performance improvement, tables yield stable results. Semantically equivalent requests (e.g. \prompt{``Give/Provide me information''} ) occasionally yield different LLM responses. Tabular formats reduce the variability in results by structurizing the relationship between entities or attributes with schemas, increasing the robustness of the model. The results are shown in Fig.~\ref{fig:var}.

 \begin{table}[ht]
     \caption{Performance comparison between four data structuring levels . Bold and underlined fonts indicate the best and second-best performances, respectively. We observe that performance improves as more structuredness are blended. Extensive results are in Appendix~\ref{app:data structuring}.}
    \centering
    \scalebox{0.9}{
    \begin{tabular}{c|c@{\hspace{10pt}}c@{\hspace{10pt}}c@{\hspace{10pt}}c}
    \toprule
    Requests & \small \bf Natural & \small \bf Order & \small \bf Template & \small \bf Table \\
    \midrule
    \bf Retrieval & 68.3 & 69.5 & \underline{70.6} & \bf 74.0 \\
    \bf Exclusion & 43.2 & 43.9 & \underline{45.8} & \bf 50.4 \\
    \bf Revision & 12.7 & \underline{14.3} & \underline{14.3} & \bf 29.8 \\
    \midrule
    \bf Superlative &30.9 & 29.9& \bf 32.0& \underline{31.3}\\
    \bf Summation & 13.9 & 15.0 & \bf 16.5 & \underline{15.6} \\
    \midrule
    \bf Quantification & 8.05 & 7.40 & \underline{6.92} & \bf 6.28 \\
    \bottomrule
    \end{tabular}
    }
    \label{tbl:structurelevel_avg}
\end{table}
\vspace{-0.1cm}
\subsection{Impact of Data Structuring Levels}
\label{subsec:data structuring}
\vspace{-0.1cm}

We interpolate between fully tabular and text inputs and construct two additional structuring levels of text, Template-based and Order-fixed (see examples in Tbl.~\ref{tab: structuring level}). For Template-based text, we alter the attribute values for a fixed template for each entity. For example, each sentence follows the structure: \prompt{"\{Name\} is a soccer player from \{Nationality\} playing for \{Club\} with uniform number \{Number\}."} For Order-fixed text, we retain the positional ordering of attributes but paraphrase the expressions of attribute values or relationships, such as changing \textit{wearing jersey number 7} to \textit{donning uniform number 7}. Examples of each format are in Tbl.~\ref{tab: structuring level}. These formats introduce structuredness to text, which originally lacks uniform structures. We compare the average performances between different structuring levels for the \texttt{Soccer} dataset in Tbl.~\ref{tbl:structurelevel_avg} (see more results in Appendix~\ref{app:data structuring}). On average, \textit{Table} achieves the best performance, and model performance shows positive correlation with input information structuredness.  

\begin{figure*}[t]
    \centering
    \includegraphics[width=0.9\linewidth]{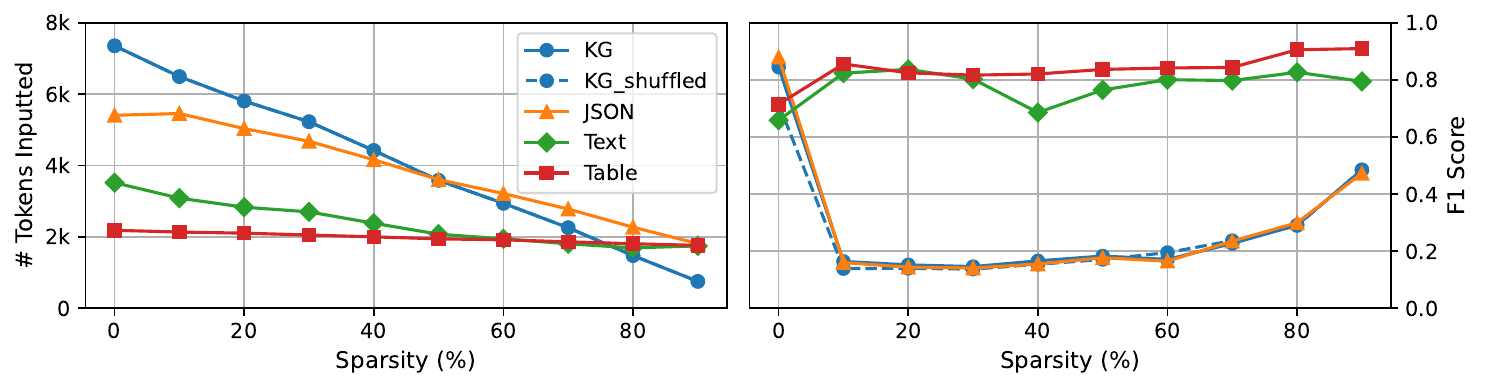}
    \vspace{-0.1cm}
    \caption{Number of input tokens (left) and F1 scores (right) of GPT-4o for \textit{Retrieval} request under varying sparsity levels for different structures (Text, JSON, KG, KG\_shuffled, and Table). Zero sparsity indicates a dense data scenario and lower token counts imply greater token efficiency. We observe that injecting semi-structured formats degrades model performance, while tables remain the most effective and token efficient.} 
    \vspace{-0.1cm}
    \label{fig:sparsity}

\end{figure*}

\vspace{-0.1cm}
\subsection{Advantages of Tabular Formats}
\label{subsec:why_tables}

As shown in the results in Sec.~\ref{subsec: main} and~\ref{subsec:data structuring}, tabular structures help improve LLMs' performance. Besides tables, semi-structured formats like JSON objects and knowledge graphs are widely adopted and are known to be helpful for LLMs~\citep{he2024does, tam2024let}. Through experiments, we find that tables offer two key benefits. 

\noindent\textbf{Sparse Data Scenarios.} When attribute values are partially missing, semi-structured formats fail to help the model as shown in Fig~\ref{fig:sparsity}. Specifically, we randomly remove $x\%$ of attributes (where $x = 10,20,\dots,90$) from the factual data source used in \textit{Retrieval} requests on the \texttt{Soccer} dataset (see results for \textit{Quantification} and \textit{Summation} requests in Appendix~\ref{app:why_table_app}). Surprisingly, across all sparsity levels, injecting JSON objects or knowledge graphs results in lower performance than providing text to the model. In contrast, tables consistently outperform other formats. Unlike semi-structured formats that become fragmented when parts of data are missing, the persistent schema of tables facilitate as a reliable anchor, preventing accuracy loss.

\noindent\textbf{Token Efficiency.} For contemporary LLMs, which are constrained by finite context windows, providing information compactly is highly beneficial. Enhanced token efficiency lowers inference costs and enables users to handle larger data. As shown in Fig.~\ref{fig:sparsity}, tabular formats mostly require less tokens than other structures or text regardless of the sparsity. Notably, in dense data scenarios, where all entities have the same set of attributes (sparsity 0\% in the figure), tables show exceptional token efficiency, while exhibiting comparable performance to semi-structured formats.

\subsection{Mechanistic Insight: Attention Focus}
\begin{table}[tbp]
    \centering
    \caption{Attention results for Llama3.1 on the \textit{Retrieval} task with the \texttt{Soccer} dataset.}
    \vspace{-0.1cm}
        \scalebox{0.9}{
        \begin{tabular}{c|c|c}
            \toprule
            $\times 10^{-2}$ & Text & Table \\
            \midrule
            \midrule
            Avg. Attention & 2.48 $\pm$ 0.30 & \textbf{5.56 $\pm$ 0.04}\\
            \bottomrule
        \end{tabular}
        }
        \vspace{-0.1cm}
        \label{tbl:attention results}
\end{table}
We take a closer look within the model to investigate why tables contribute to better performance with Llama3.1:70B, through attention analysis. For \textit{Retrieval} requests on the \texttt{Soccer} dataset, we compare the aggregated attention values across all heads and layers for table elements (schema and row) against the value for the corresponding textual sentences, relative to the request. Tabular formats receive 2.24 times the attention weights of textual counterparts, as demonstrated in Tbl.~\ref{tbl:attention results}, with lower standard deviation, indicating higher stability. These findings suggest that tables assist models to focus on the relevant contents, resulting in more accurate and reliable predictions.


    

\vspace{-0.1cm}
\subsection{Robustness to Request Complexity}
\label{exp: cond num}

To evaluate scalability under increased request complexity, we vary the number of conditions for \textit{Retrieval} requests on the \texttt{Soccer} dataset. Note that we concatenate the conditions with an ``or'' operator as the intersection will likely be empty when the number of conditions increases. We show the results of three models in Fig.~\ref{fig:condition_portion} (others are shown in Appendix~\ref{app: cond num}). While the performance decreases as the number of conditions increases, tabular structures consistently benefit LLMs compared to text regardless of the number of conditions. 

\vspace{-0.1cm}
\subsection{Resilience to Incomplete Structuring}
\label{subsec:struct_portion}

We vary the proportion of the input presented in tabular format using a mixing coefficient \(\alpha\in\{0,0.25,0.5,1.0\}\), where $\alpha$ denotes the fraction of data in tabular form and $1-\alpha$ the fraction in natural text. This setup allows us to explore whether tabular formats remain beneficial even when the whole data cannot be represented as tables. This setup addresses practical settings where information is often only partially structured, leaving portions of the data in a non-tabular format. For example, $\alpha=0.25$ means $25\%$ of data is in a tabular format, while the remaining $75\%$ is presented as natural text. We conduct the experiment for \textit{Retrieval} requests on the \texttt{Soccer} dataset, with the results for three models shown in Fig.~\ref{fig:condition_portion} (other results are in Appendix~\ref{app:struct_portion}). As illustrated in the figure, even structuring a quarter of the information improves performance for most models (3.4pp improvement on average), implying that small incorporation of structures can also benefit LLMs. 

\begin{figure}[t]
    \includegraphics[width=\linewidth]{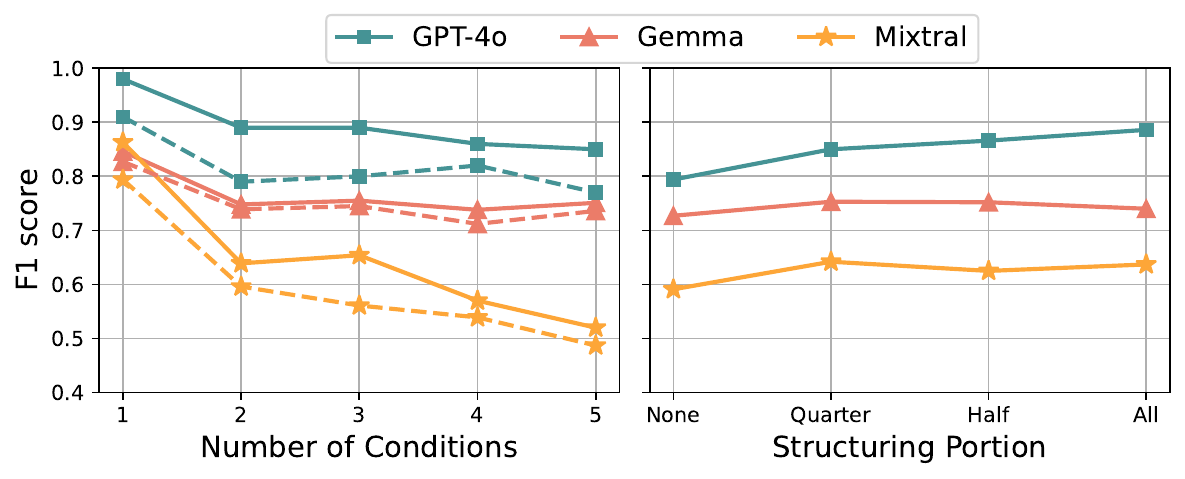}
    \captionsetup{skip=2pt}
    \caption{LLM performance across varying number of conditions (Left) and structuring portions (i.e. fraction of data in tabular form with the remainder as text) (Right). The dotted and solid lines are the results for \textit{Text} and \textit{Table}, respectively.} 
    \label{fig:condition_portion}
\end{figure}

\begin{figure}[t]
    \centering
    \includegraphics[width=0.9\linewidth]{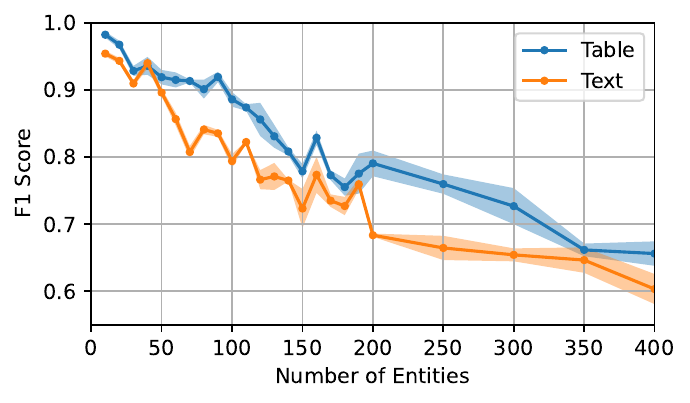}
    \captionsetup{skip=2pt}
    \caption{LLM performance across varying context length (i.e. number of entities inputted). The shaded bands indicate ±1 standard deviation over templates.}
    \label{fig:length}
\end{figure}



\jo{
\vspace{-0.1cm}
\subsection{Scalability to Context Length}
We vary the amount of input data provided to GPT-4o for \textit{Retrieval} requests on the Soccer dataset. We investigate whether the advantages of tabular representations persist as the context scale varies. As illustrated in Fig.~\ref{fig:length}, we observe a a constant decline in performance as the input data size grows, while tabular structures consistently outperform textual formats, providing an average improvement of 5.30 pp. This performance boost implies that tables can benefit LLMs in long‐context scenarios, which are known to be challenging for modern LLMs~\cite{liu2023lost, li2024long}.}


\section{Discussion}
\label{subsec:ext}

\subsection{Generalizing Tabular Benefits to Unstructured Contexts}
Informational requests often rely on factual attributes that can be captured within a tabular schema. To evaluate whether models can structurize knowledge into tables, we prompt LLMs to transform unstructured text into tabular representations across three datasets. We find that LLMs are able to correctly fill in 98.6\% of the rows and columns on average, even when the columns are not provided to the models (e.g., ``\prompt{Convert the given text to a table}''). This ability of LLMs to organize  information suggests that any factual source capable of textual conversion (e.g., KG) can utilize the benefits of tabular formatting for various tasks.

\subsection{Extensibility of Request Types}
To demonstrate the broader applicability and completeness of our framework, we evaluate three additional request types based on formal relational logic: Existence (logical quantification), Projection (attribute-specific extraction), and Difference Conditions (set negation). These tasks extend the main requests to cover a wider spectrum of complex information seeking user needs. Across these scenarios, tabular representations consistently show higher performance and robustness over textual formats. Detailed formalizations and results are provided in Appendix \ref{subsec:expanding_to_sql}.

\section{Related Work} \label{sec: related}

\subsection{LLM with Complex Requests.}
Recently, many works have been conducted on how difficult LLMs understand and generate accurate outputs for complex information retrieval or data manipulation~\citep{barke2024solvingdatacentrictasksusing,he2024can,xiong2020answering,zhang2024cfbench}. 
The most relevant work is presented by \citet{he2024complex}, which also propose a method to enhance LLM performance on complex instructions. However, their focus is primarily on how to obtain and utilize effective training data, while we thoroughly demonstrate the impact of data structuredness, especially tables.



\subsection{LLM with Tables.}
Tabular structure is an organized format that contains large amounts of information. This systematic characteristic makes tabular data essential for many applications~\citep{deng2022recent,deng2022turl, chen2020hitter,chen2020logical}. Prior studies explore different methods to utilize, represent and encode tabular data. \citet{jin2023tab} show providing zero-shot chain-of-thought (CoT) prompts~\citep{kojima2022large}(adding ``Let's think step by step'' after the prompt) in a tabular form (``|step|event|answer'') improves LLM performances on reasoning tasks.
\citet{hegselmann2023tabllm} report that text outperforms list enumeration formats for prediction or classification tasks. 
\citet{wang2024chain} suggest a multi-step tabular reasoning approach with table evolution to improve table understanding. \citet{deng2024tables} systemically evaluate how different text-based or image-based prompt methods affect LLMs’ performances on table-related tasks. In contrast, our work shows that tabular formats yield higher performance, robustness, and token-efficiency for prevalent requests on factual data and our findings integrates seamlessly with  prompt‑optimization and CoT techniques.




\section{Conclusion} \label{sec: conclusion}
We show that providing tabular structures enhance LLM performance compared to other formats for prevalent requests on factual data, which are common in real-world user interactions and current LLM benchmarks. The key intuition is to align with human cognitive preferences, where humans benefit from organizing information with structures, representatively tables, when dealing with complex tasks. We show that tabular formats significantly improves LLM performance, reduces output variance, and the most token efficient. Moreover, through comprehensive ablation studies, we show the generalizability and completeness of our approach. We present possible future extensions, highlighting the potential of structured representations to address intrinsic limitations of text. We highlight the overlooked potential of tables, establishing a framework for future use of LLMs.

\section*{Limitations}
\label{sec: discussion}

\vspace{-0.1cm}
\paragraph{Extensibility to Non-Tabular Data Sources.} Other popular factual storage formats such as knowledge graphs offer better flexibility that exceeds the expressive power of two-dimensional tabular formats (e.g., nested or hierarchical graphical relationships). In this paper, we focus on requests where the needed factual attributes rarely depends on higher-order relationships. For further questions that require resolving complex hiearchial relationships, different structures might benefit more. Nonetheless, in this paper, we show that more structuredness helps that hybrid approaches (i.e. tables along with text or knowledge graphs) are still viable.

\paragraph{Beyond Factual Requests.} We mainly focus on requests that targets to analyze, retrieve, or manipulate factual data in this paper. However, tables have the potential to benefit models across a wide range of prompts extending beyond request types dealt in our work. For example,~\citet{jin2023tab} demonstrate that even Zero-Shot CoT that uses a table-structured input like \prompt{|step|event|answer|} instead of the conventional \prompt{Let's think step by step} can benefit from tabular cues for reasoning tasks. Moreover,~\citet{he2024does} show that structuring prompt formats, for instance to JSON formats, help increase model performance and stability. Hence, tables have the potential to improve performance beyond factual requests across various tasks.

\bibliography{main}

\clearpage
\appendix

\section{Related Works about LLM-assisted Table Management} Recently, approaches to utilize LLM for effectively managing data with tables are also suggested. \citet{sun2023sql,hong2024knowledge,hong2024next} propose techniques for generating SQL queries using LLMs and provide insights into improving the interaction between user queries and database schemas.
\citet{arora2023language,wu2021text,zhang2024tablellm} investigate to extract tables from diverse data sources including semi-structured tables, texts, and images. 
Our goal is to improve model performance on prevalent requests on factual data, which LLMs often fail to answer correctly, rather than utilizing LLMs as table management tools.

\section{Models and Hyperparameter Details}
\label{app: model details}

GPT models are run through the Azure OpenAI API. Gemini and Claude models are run through the Google and Anthropic APIs, respectively. The open-sourced models, Mixtral-8x22B, Llama-3.1:70B, and Gemma2-27B, are run on 16 A100 GPUs without parallelism, respectively. All models' temperature parameters are set to zero to control randomness.

\begin{table}[t]
    \caption{Extensive results of Fig.~\ref{fig:condition_portion} across the varying number of conditions. The numbers are F1 scores, scaled into a percentage scale.}
    \label{tbl:condition_num}
    \centering
    \scalebox{0.9}{
    \begin{tabular}{c|c|c@{\hspace{10pt}}c@{\hspace{10pt}}c@{\hspace{10pt}}c@{\hspace{10pt}}c}
    \toprule
     \multirow{2}{*}{Model} & Data & \multicolumn{5}{c}{Number of Conditions}\\
     & Type & 1 & 2  & 3  & 4  & 5\\
    \midrule
    
     \multirow{2}{*}{GPT-3.5} & Text & 56.1 & 18.3 & 17.3 & 17.8 & 17.6\\ 
     & Table & \textbf{79.7}& \textbf{26.8} & \textbf{28.2} &\textbf{29.9}&\textbf{27.5}  \\
     \midrule
     \multirow{2}{*}{GPT-4} & Text & 92.4 & 85.7 & 84.5 & 83.0 & 79.4\\ 
     & Table & \textbf{94.6}& \textbf{93.6}& \textbf{90.3} &\textbf{88.6}&\textbf{86.5}  \\
     \midrule
     \multirow{2}{*}{GPT-4o} & Text & 91.3 & 79.9 & 80.5 & 81.9 & 76.7\\ 
     & Table & \textbf{97.7}& \textbf{89.6}& \textbf{89.3} &\textbf{86.2}&\textbf{85.1}  \\
     \midrule
     \multirow{2}{*}{Gemini} & Text & \textbf{96.7} & 75.9 & \textbf{82.6} & 77.6 & 78.4\\ 
     & Table & 92.9&\textbf{88.0} & 81.9 &\textbf{79.2}&\textbf{80.3}  \\
    \midrule
     \multirow{2}{*}{Claude} & Text & 80.4 & \textbf{76.9}& 67.6 & 66.8 & 63.4\\ 
     & Table & \textbf{87.9}& 73.9& \textbf{69.3} &\textbf{67.7}&\textbf{68.2}  \\
     \midrule
     \multirow{2}{*}{Mixtral} & Text & 79.4 & 59.6& 56.1 & 53.9 & 48.7\\ 
     & Table & \textbf{86.3} & \bf 63.9 & \textbf{65.4} &\textbf{57.0}& \textbf{52.0} \\
     \midrule
     \multirow{2}{*}{Llama} & Text & 86.7 & 78.6 & \textbf{76.8} & \textbf{73.7} & \textbf{74.3}\\ 
     & Table & \textbf{90.3} & \textbf{83.1} & 75.0& 70.4& 70.0 \\
     \midrule
     \multirow{2}{*}{Gemma} & Text & 82.7 & 73.9\ & 74.5  & 71.2  & 73.6 \\ 
     & Table & \textbf{84.6} & \bf 74.8 & \textbf{75.5}  &\textbf{73.8}& \textbf{75.1}\\
    \bottomrule
    \end{tabular}
    }
  \end{table}

\section{Experiment Details}

\subsection{Table Representations in LLMs}
We utilize the delimiter ``|'' for table formatting introduced by~\citet{jin2023tab}. They demonstrated that proper formatting is crucial for LLMs to accurately interpret tabular structures, while alternative delimiters, such as ``,'', lead to failure in capturing the inherent structure of tables.

\subsection{Advantages of LLM for Informational Requests}
\label{app:motivation}
In this paper, we evaluate the performance of LLMs on informational or factual tasks when different structures are injected as background information or knowledge. Although other tools such as SQL, SPARQL, or pandas are known to be good tools for these tasks, LLMs offer distinct advantages for everyday users. Here, we give three representative examples.

\paragraph{Accessibility for Non-experts.} As mentioned in Sec.~\ref{sec: intro}, LLMs are being widely used as tools in daily life, being alternatives for search engines. Ordinary users lack database or computer science expertise, thus usually provide these requests in a natural language form. Hence, reliable and robust LLM outputs for these request build trust and enable normal users to easily perform these tasks. 

\paragraph{Utilizing LLM Capabilities.} Beyond requests that are mapped to SQL or other formal query languages, LLMs allow users to perform diverse tasks with the same data. For instance, users can pose follow-up questions such as \prompt{``Which of the players scored hat-tricks?''} or \prompt{``Which players are going to perform well next season?''}, which cannot be expressed in structural languages. For such questions, it is essential that LLMs interpret the underlying factual data, a capability we rigorously evaluate through various scenarios in this paper. Hence, ensuring correctness for these factual requests as a first step is crucial to avoid snowball effects~\cite{zhang2023language,oh2024erbench}, compounding accumulations of small errors that often lead LLMs to hallucinate.

\subsection{Condition Generation.}
\label{app: cond}
con
Among all attributes in the dataset, we randomly select an attribute and the corresponding value for generating each condition. For example, if the \textit{nationality} attribute is chosen, one value is randomly selected among the unique values of \textit{nationality} in the dataset, such as \textit{Spain}, \textit{England}, or \textit{Portugal}. Plus, we vary the condition types as well. For the \texttt{Soccer} dataset, conditions are based on exact equality, whereas for the \texttt{Movie} and \texttt{PII} datasets, we use inequality and partial equality. For example, \prompt{"Nationality is Argentina"} is based on equality, while \prompt{"Rating is higher than 3.0"} and \prompt{"Domain is gmail"} are based on inequality and partial equality, respectively. These strategies ensure the generation of diverse and unbiased set of questions. Additionally, we verify that the prompted table and text mostly have entities that satisfy the generated conditions. This allows us to avoid edge cases where no entities satisfy the conditions. We generate two conditions and concatenate these conditions with one of two logical operators, ``and'' or ``or''.

\subsection{Dataset Information}
\label{app: dataset}

Continuing from Sec.~\ref{subsec:llmsdatasets}, we use three datasets presenting different domains. The \texttt{Soccer} and \texttt{Movie} datasets are tabular datasets; and the \texttt{PII} dataset consists of AI-generated text and the table extracted from the text. We select 100 entities for each dataset in the experiments.

\begin{itemize}
    \item \textbf{Soccer~\cite{fifa20}:} We use a relation with the attributes  \textit{player name}, \textit{club}, \textit{jersey number}, \textit{nationality}, \textit{league}, and \textit{preferred foot}.
    \item \textbf{Movie~\cite{imdb}:} We use a relation with the attributes \textit{movie title}, \textit{director name}, \textit{movie length}, \textit{actor name}, \textit{released year}, \textit{movie}, and \textit{rating}. 
    \item \textbf{PII~\cite{pii}:} This dataset contains AI-generated texts and several information extracted from those texts. The extracted relation has the attributes \textit{name}, \textit{email}, \textit{phone number}, \textit{job}, \textit{address}, \textit{hobby}, and \textit{job experience years}.
\end{itemize}

All datasets are publicly available allowing any usages on Kaggle with CC0: Public Domain license for the \texttt{Soccer} dataset, Public Domain license for the \texttt{Movie} dataset, and Apache 2.0 license for the \texttt{PII} dataset. All datasets are mostly in English with no privacy or ethical concerns.

\begin{table*}[t]
    \centering
    \caption{Textual representation of semi-structured formats. ``KG'' refers to knowledge graphs. }
    \scalebox{1.0}{
    \begin{tabular}{c|l}
    \toprule
      \multirow{2}{*}{Format}&  \multirow{2}{*}{Example}\\
     & \\
    \midrule
 \multirow{2}{*}{KG} & (\textbf{Ronaldo}, Number, \textbf{7}), (\textbf{Ronaldo}, Nationality, \textbf{Portugal}), (\textbf{Ronaldo}, Club, \textbf{Juventus}),\\
    &   (\textbf{Messi}, Number, \textbf{10}), (\textbf{Messi}, Nationality, \textbf{Argentina}), (\textbf{Messi}, Club, \textbf{Barcelona}) \\
    \midrule
    \multirow{2}{*}{JSON}  & [\{Name: \textbf{Ronaldo}, Number: \textbf{7}, Nationality: \textbf{Portugal}, Club: \textbf{Juventus}\}, \\
    & \{Name: \textbf{Messi}, Number: \textbf{10}, Nationality: \textbf{Argentina}, Club: \textbf{Barcelona}\}] \\
    \bottomrule
    \end{tabular}
    }
    \label{tab: semi-structure_representation}
\end{table*}

\begin{table*}[t]
    \caption{Performance comparison on the \textit{multi-turn} scenario on the \textit{Soccer} dataset.}
    \label{tbl:multiturnperformance}
    \centering
    \scalebox{0.9}{
    \begin{tabular}{c|c|c@{\hspace{8pt}}c@{\hspace{8pt}}c@{\hspace{8pt}} c@{\hspace{8pt}}c@{\hspace{8pt}}c@{\hspace{8pt}}c|c}
    \toprule
    Request Type & Data Type & GPT-3.5 & GPT-4 & GPT-4o & Gemini & Mixtral & Llama & Gemma & Avg. \\
    \midrule
    \midrule
    \bf Retrieval & Text & 17.9 & 64.2 & 77.6 & 66.9 & 47.3 & 63.6 & 45.9 & 54.8 \\ 
    (F1 score, \%) & Table & \textbf{34.4} & \textbf{86.9} & \textbf{82.7} & \textbf{76.0} & \textbf{62.1} & \textbf{73.8} & \textbf{66.9} & \textbf{69.0} \\
    \midrule
    \bf Exclusion & Text & 31.2 & 50.4 & 55.7 & \textbf{39.4} & \textbf{23.2} & \bf 53.9 & \textbf{23.3} & 39.6 \\ 
    (F1 score, \%) & Table & \textbf{50.2} & \textbf{73.9} & \textbf{71.5} & 37.5 & 12.5 &  44.1& 17.2 & \textbf{43.8} \\
    \midrule
    \bf Revision & Text & 1.2 & 25.7 & 29.8 & 17.4 & 1.2 & \bf 12.5 & 1.2 & 12.7 \\ 
    (F1 score, \%) & Table & \bf 16.3 & \bf 51.8 & \bf 63.3 & \bf 26.9 & \bf 5.7 & 11.0 & \bf 5.4  & \bf 25.8 \\
    \midrule
    \midrule
    \bf Superlative & Text & 11.7 & 40.8 & 28.5 & 42.2 & 14.5 & 31.3 & \textbf{19.8} & 27.0 \\ 
    (Accuracy, \%) & Table & \textbf{17.3} & \textbf{47.0} & \textbf{33.5} & \textbf{47.3} & \textbf{20.8} & \textbf{31.7} & 17.0 & \textbf{30.7} \\
    \midrule
    \bf Summation & Text & \textbf{5.8} & \textbf{18.5} & 10.5 & 22.3 & 7.7 & 9.0 & \textbf{9.7} & 11.9 \\ 
    (Accuracy, \%) & Table & 4.5 & 17.7 & \textbf{11.3} & \textbf{26.2} & \textbf{8.8} & \textbf{9.2} & 5.7 & 11.9 \\
    \midrule
    \midrule
    \bf Quantification & Text & \textbf{13.35} & 12.54 & 5.31 & 14.46 & \textbf{11.26} & 9.85 & 9.91 & 10.95 \\
    (Abs. Diff. $\downarrow$) & Table & 13.41 & \textbf{10.19} & \textbf{4.30} & \textbf{11.34} & 11.95 & \textbf{8.33} & \textbf{8.54} & \textbf{9.72} \\
    \bottomrule
    \end{tabular}
        }
\end{table*}

\subsection{Dataset Manipulation for Fair Comparison}
\label{app: fair comparison}
Continuing from Sec.~\ref{subsec: main}, intuitively, it seems more challenging to extract information and provide appropriate analyses from natural texts than from tables. Therefore, we adopt two approaches to facilitate the construction of text-table pairs for a fair comparison. The first method is called a \textit{text-based} construction, which involves removing extraneous information from the natural text. After extracting the information and constructing tables, irrelevant sentences are eliminated. This method is straightforward and intuitive, but has limitations when it comes to accurately analyzing experimental results. (1) The first limitation is that the difficulty of extracting and analyzing information can vary depending on the complexity of each sentence. (2) The second limitation is that irrelevant information may remain if it appears within the same sentence as essential information. 

To address these limitations, we also employ a reverse approach called a \textit{table-based} construction, which constructs natural sentences based on the attributes in the table. By using this method, we can generate multiple levels of text that include exactly the same amount of information as the table, as described in Tbl.~\ref{tab: structuring level}. 

As a result, we generate text-table pairs using the \textit{table-based} method for the \texttt{Soccer} and \texttt{Movie} datasets and \textit{text-based} method for the \texttt{PII} dataset.

\subsection{Representation of Semi-structured Formats}
\label{app:representation_semi_struct}

Semi‐structured formats, which are also widely used for storing factual information, lie between text and tables, annotating data with explicit keys, while permitting variable ordering. The exemplar representation of semi-structured formats are shown in Tbl.~\ref{tab: semi-structure_representation}. Although JSON and knowledge graphs have variable attribute ordering, we fix the attribute orders for both in our experiments, which enhances structuredness.

\section{Extended Results}
\subsection{More Results from Sec.~\ref{subsec: main}}
\label{app:more_main_res}
We present the results of the \textit{multi-turn} and \textit{pre-instruction} scenario in Tbl.~\ref{tbl:multiturnperformance} and Tbl.~\ref{tbl:preinstruction_res}, respectively.
In the \textit{multi-turn} scenario, the model structures the factual data in text or table format in the first turn and then answers the subsequent data-analytics request based on that information. As shown in Tbl.~\ref{tbl:multiturnperformance}, tabular formats consistently outperforms the text baseline across all request types. 
In the \textit{pre-instruction} scenario, we prompt the model to internally structure textual information into a tabular format. As shown in Tbl.~\ref{tbl:preinstruction_res}, although reasoning models achieve relatively high performance, the instruction yields additional gains, still having room for improvement. This result implies that LLMs not only benefit from injecting structuredness, but also from being nudged to construct tabular structures internally in their intermediate outputs. 

\paragraph{Generalizability to Smaller LLMs.} We originally test state-of-the-art models with bigger model sizes to ensure that the model has the capability to understand tabular structures and correctly perform different tasks, following ~\citet{jin2023tab}. We conduct additional experiments on Llama 3.2-1B, 3B, and Llama 3.1-8B for the \textit{single-turn} scenario, as shown in Tbl.~\ref{tbl:smaller_models}. The results show that the benefit of tabular input becomes more evident as model capability increases. For instance, the table format improves retrieval and exlcusion accuracy for Llama 3.1-8B, but offers limited or even negative gains for smaller models. Interestingly, within the same model, relative performance advantage between text and tables diverge on certain tasks (e.g., Quantification and Summation), suggesting that tabular advantages depend not only on model size but also on the model’s ability for that task. These findings indicate that the effect of structured representation is not strictly model- or task-specific, but rather contingent on the model’s overall competence level. In other words, tabular structures are beneficial once the model reaches sufficient model capability, while smaller models may struggle to utilize such structure effectively.

\paragraph{Ablation on Non-Reasoning Models with \textit{Pre-instruction}.} To test whether \textit{pre-instruction} helps for models without test-time compute, we perform additional experiments with a non-reasoning model (GPT-4o) as shown in Tbl.~\ref{tbl:preins_gpt4o}. Unlike the results shown on reasoning models, this additional instruction does not yield consistent improvements and in most tasks, reuslts in performance degradation. This finding suggests that the benefit of \textit{pre-instruction} is not a product of simple prompt engineering, but rather from the actual structurization that the model goes through in their intermediate output steps. This result strengthens our claim that structures, especially tabular structures, help LLM performance.

\begin{table}[ht]
    \caption{Performance comparison for GPT-4o on the \textit{Soccer} dataset to evaluate the impact of \textit{pre-instruction} on a non-reasoning model.}
    \label{tbl:preins_gpt4o}
    \centering
    \setlength{\tabcolsep}{4pt} 
    \scalebox{0.85}{ 
    \begin{tabular}{c|ccc}
    \toprule
    \multirow{2}{*}{Request Type} & \multirow{2}{*}{Table}& \multirow{2}{*}{Text} & Pre-ins
    \\ &&&+ Text\\
    \midrule
    \midrule
    \bf Retrieval & \multirow{2}{*}{\bf 88.6} & \multirow{2}{*}{79.4} & \multirow{2}{*}{75.6} \\
    (F1 score, \%)&&&\\
    \midrule
    \bf Exclusion & \multirow{2}{*}{\bf 70.6} & \multirow{2}{*}{62.9} & \multirow{2}{*}{54.2} \\
    (F1 score, \%)&&&\\
    \midrule
    \midrule
    \bf Summation & \multirow{2}{*}{\bf 13.0} & \multirow{2}{*}{13.2} & \multirow{2}{*}{21.7}\\
    (Accuracy, \%)&&& \\
    \midrule
        \bf Quantification &\multirow{2}{*}{\bf 3.9} & \multirow{2}{*}{4.3} & \multirow{2}{*}{5.3} \\
    (Abs. Diff., $\downarrow$)&&& \\
    \bottomrule
    \end{tabular}
    }
\end{table}

\begin{table}[ht]
    \caption{Performance comparison of smaller models on the \textit{Soccer} dataset.}
    \label{tbl:smaller_models}
    \centering
    \setlength{\tabcolsep}{4pt} 
    \scalebox{0.9}{ 
    \begin{tabular}{c|c|ccc}
    \toprule
    \shortstack{Request \\ Type} & \shortstack{Data \\ Type} & \shortstack{Llama \\ 3.1-8B} & \shortstack{Llama \\ 3.2-3B} & \shortstack{Llama \\ 3.2-1B} \\
    \midrule
    \midrule
    \bf Retrieval & Text & 55.4 & 29.0 & \bf 22.9 \\
    (F1 score, \%) & Table & \bf 60.2 & \bf 41.1 & 14.7 \\
    \midrule
    \bf Exclusion & Text & 35.4 & \bf 26.8 & \bf 26.5 \\
    (F1 score, \%) & Table & \bf 42.9 & 19.9 & 25.2 \\
    \midrule
    \midrule
    \bf Summation & Text & \bf 8.8 & 6.0 & \bf 4.8 \\
    (Accuracy, \%) & Table & 8.0 & \bf 6.5 & 1.5 \\
    \midrule
    \bf Quantification & Text & \bf 0.9 & 15.1 & 14.8 \\
    (Abs. Diff. $\downarrow$) & Table & 11.7 & \bf 13.4 & 14.8 \\
    \bottomrule
    \end{tabular}
    }
\end{table}

\begin{table}[ht]
   \caption{Performance comparison for the \textit{pre-instruction} scenario on the \textit{Soccer} dataset. The models o3-mini and 2.5-flash correspond to GPT-o3-mini and Gemini-2.5-flash (thinking budget: 20000), respectively.}
    \label{tbl:preinstruction_res}
    \centering
    \scalebox{0.9}{
    \begin{tabular}{c|c|cc@{\hspace{8pt}}}
    \toprule
    Request Type & Data Type & o3-mini & 2.5-flash \\
    \midrule
    \midrule
    \bf Retrieval & Text & 97.4 & 99.3 \\
    (F1 score, \%) & Table & \textbf{97.7} & \textbf{99.4} \\
    \midrule
    \bf Exclusion & Text & 82.9 & 88.6 \\
    (F1 score, \%) & Table & \textbf{84.8} & \textbf{88.7} \\
    \midrule
    \bf Revision & Text & \bf 43.1 & 59.5 \\
    (F1 score, \%) & Table & 38.4 & \bf 60.3 \\
    \midrule
    \midrule
    \bf Superlative & Text & 86.0 & 95.3 \\
    (Accuracy, \%) & Table & \bf 88.2 & \bf 95.8 \\
    \midrule
    \bf Summation & Text & \bf 81.7 & \bf 82.5 \\
    (Accuracy, \%) & Table & 80.8 & 80.2 \\
    \midrule
    \midrule
    \bf Quantification & Text & 0.32 & \bf 0.17 \\
    (Abs. Diff. $\downarrow$) & Table & \bf 0.26 & 0.32 \\
    \bottomrule
    \end{tabular}
    } 
\end{table}

\begin{figure*}[t]
    \centering
    \includegraphics[width=\linewidth]{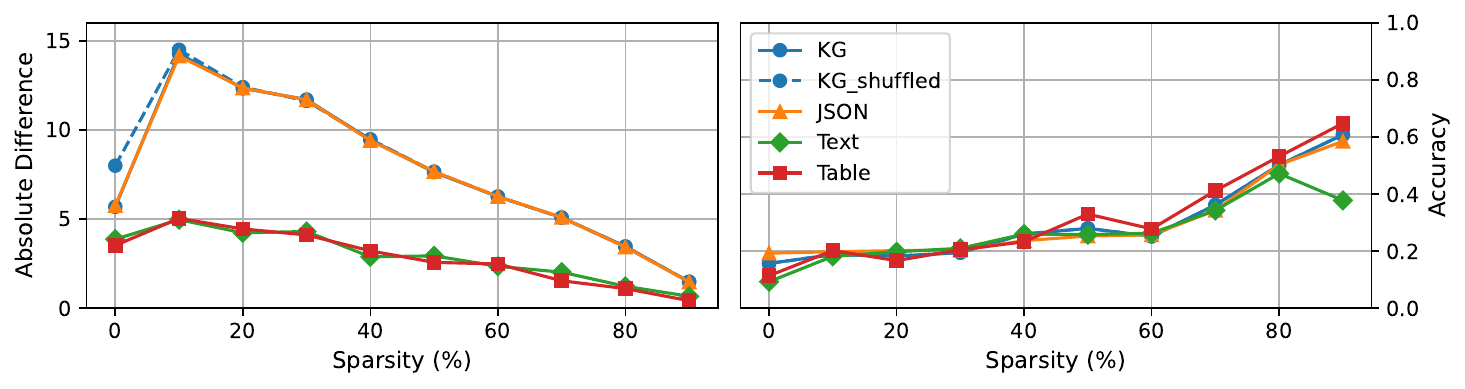}
    \caption{Absolute difference for \textit{Quantification} request (left) and accuracy of GPT-4o for \textit{Summation} request (right) under varying sparsity levels for different structures (Text, JSON, KG, KG\_shuffled, and Table). Zero sparsity indicates a dense data scenario. Note that lower absolute difference and higher accuracy indicate better performance.} 
    \label{fig:sparsity_app}
\end{figure*}

\subsection{More Results from Sec.~\ref{subsec:data structuring}}
\label{app:data structuring}
We present the results between different structuring levels in Tbl.~\ref{tbl:structurelevel_performance}. Bold and underlined fonts indicate the best and second-best performances, respectively. In general, models show better performances when more structures are blended into text, while tables are the most effective.

\subsection{More Results from Sec.~\ref{subsec:why_tables}}
\label{app:why_table_app}
We present the results in Fig.~\ref{fig:sparsity_app} across five input formats for \textit{Quantification} and \textit{Summation} requests on different sparsity levels, as we mask out x\% of the attributes. Note that for
KG and JSON, while we fix the attribute orders
in our experiments, these formats typically do not require consistent orders of attributes across entities. Moreover, since KGs do not have guarantees for entities to come in order, we also compare the scenario where the sequence of graph triples, (Subject, Predicate, Object), is randomized, denoted as ``KG\_shuffled''.
Consistent with our findings for the \textit{Retrieval} request, semi-structured formats (JSON and KG) lead to performance degradation when data are sparse, while tables remain relatively stable and most effective. Interestingly, the performance for \textit{Quantification} and \textit{Summation} requests increases as data become sparser. This improvement occurs because with higher sparsity, the likelihood of entities satisfying certain conditions decreases, leading to more trivial or smaller answers (e.g., zero) with a reduced answer candidate space. For instance, for \textit{Quantification} requests, when the model predicts small numbers, the ground-truth answer is also typically small, resulting in a relatively small absolute error compared to denser settings.

\subsection{More Results from Sec.~\ref{exp: cond num}}
\label{app: cond num}
We present the results of varying number of conditions in Tbl.~\ref{tbl:condition_num}. In most cases, tabular formats outperforms the baseline (natural text) regardless of the number of conditions. We observe that the F1 scores decline as more conditions are added, which reflect the increased task complexity, while the advantage of inputting tabular structures to LLMs persists. Thus, we can conclude that the framework remains effective even when the request complexity increases.

\subsection{More Results from Sec.~\ref{subsec:struct_portion}}
\label{app:struct_portion}
We vary the proportion of the input data structured as tabular formats (with the remainder as text) and report the results in Tbl.~\ref{tbl:portion}. Most models' performances improves with increased portion of structures. Notably, Gemini shows significant performance gains even when only a quarter of the data is structured as tables (up to 14pp). This result shows that the framework is effective even when all background information cannot be structured into tables.

\subsection{More Results from Sec.~\ref{subsec:ext}}
\label{subsec:expanding_to_sql}
\subsubsection{Expanding Requests}
\label{subsubsec:expand_db_texxt}
The requests that handles factual data often inherently rely on principles of database query languages such as relational logic or database query languages. As shown in  Tbl.~\ref{tbl:promptrelationalalgebra}, requests tested in this paper can be converted into a database query language. Note that this conversion can also be done in the opposite direction (i.e. database query language to request), allowing diversification of request types. In the paper, we test for the most  widely used request types in (Sec.~\ref{subsec: main}) and in this section we show that tables consistently outperforms the textual baseline even for extended requests.

\paragraph{Existence.} \citet{oh2024erbench} demonstrate that factual hallucinations in an LLM's rationale can be automatically evaluated with relational databases along with its integrity constraints. Providing tables to LLMs improves performance on requests that require verifying the existence of certain entities using an existence quantifier in relational calculus, e.g., \prompt{Is there a soccer player from South Korea who played for Tottenham Spurs in 2019?}. In accordance with prior work, we report rationale accuracy (\textbf{R}) as our results.

\paragraph{Projection.} Beyond \textit{Retrieval} requests, one may want to extract an entity's attributes and properties, highly related to the projection operation in relational algebra. For example, one might wish to retrieve soccer players' names along with their jersey numbers that satisfy specific conditions.

\paragraph{Difference Condition.} Going beyond the ``and ($\land$)'' and ``or ($\vee$)'' logical operators for condition generation, ``set difference ($\setminus$)'' can be used to merge two elements in relational algebra operations from set theory~\citep{relationalcalc}. $A\setminus B$ can be expressed as $A\land \neg B$, so an example of a condition could be \prompt{``preferred foot is left and nationality is not Argentina''}. 



\subsubsection{Results}
\paragraph{Existence.}
\label{app:erbench}
As shown in Tbl.~\ref{tbl:promptrelationalalgebra}, questions proposed in \citet{oh2024erbench} can be also expressed with an existential operator in relational calculus. We test whether providing tables instead of text improves the model performance on the \textit{Soccer} dataset shown in Tbl.~\ref{tbl:erbench}. The red numbers in the parenthesis in the table indicate the performance difference between the original and negated request template (BN(Y) and BN(N) in \citet{oh2024erbench}, respectively), where lower values indicate higher robustness and consistency. Tabular formats outperforms the baseline (text) for most models, 6.5pp and 6.9pp on average for original and negated requests, respectively and 4.1pp reduction in variance between the original and negated requests.

\begin{figure*}[t]
    \centering
    \includegraphics[width=0.99\linewidth]{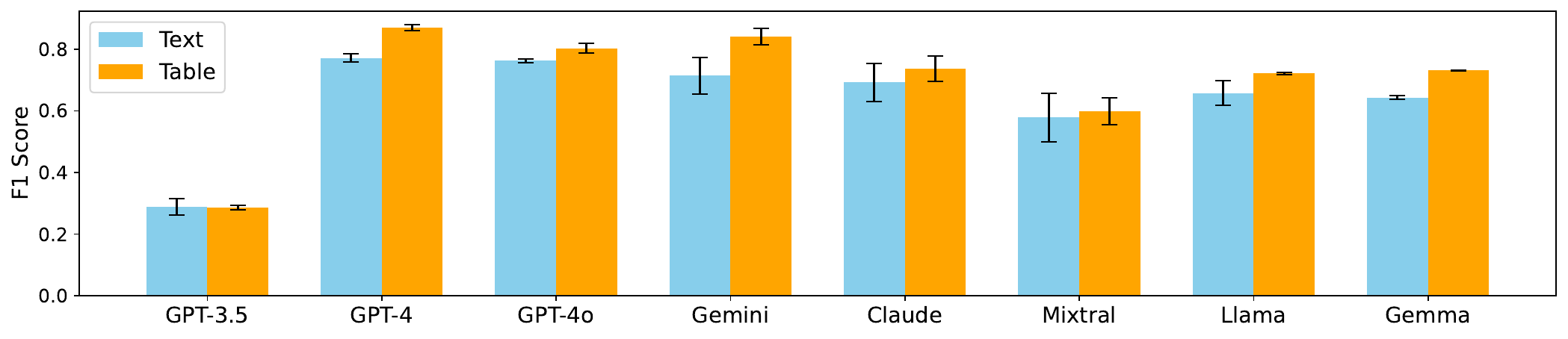}
    \caption{Results of Projection requests. Error bars indicate the variability of LLM responses across three semantically equivalent questions. We observe that a tabular structure consistently improves F1 scores (performance) and robustness.}
    \label{fig:projection}
\end{figure*}

\paragraph{Projection.}
\label{app:projection}
Consistent to the results for existence request types, LLMs show better performance and robustness when provided with tables compared to text as shown in Fig.~\ref{fig:projection}. GPT-4 and Gemini show large amounts of performance improvements compared to using natural text. On the other hand, for GPT-3.5, Claude, and Mixtral, the enhancements are relatively smaller. However, these LLMs exhibit much lower variance (error bars) across different instruction templates, indicating greater stability.

\paragraph{Difference Conditions.}
\label{app:minusoperator}
We show that the requests can be extended also in terms of conditions introducing an additional logical operator, ``$\setminus$'' (set difference). Tbl.~\ref{tbl:minusoperator} shows the experimental results for three request types for the \textit{Soccer} dataset. Tabular formats outperforms the baseline (text) with an average relative increase of 26.45\% (4.83pp).

\subsection{Case Study of Multi-turn Requests}
\label{subsec:casestudy_multiturn}
Tbl.~\ref{tbl:casestudy_multiturn} shows a common occurrence of cases in a \textit{multi-turn} scenario for Gemini. With the gold answer being zero for the corresponding question (\textit{Quantification} request), the model successfully gives a correct answer when the model outputs its factual data as a tabular form. However, when the model organizes the information as text, the model fails to provide an appropriate answer, claiming that a database would be required for an efficient computation. This result implies that structures benefit LLMs in terms of effectiveness and reliability.

\begin{table*}[t]
    \caption{Performance comparison of different structured levels on the \textit{Soccer} dataset.}
    \label{tbl:structurelevel_performance}
    \centering
    \scalebox{0.9}{
        \begin{tabular}{c|c|c@{\hspace{8pt}}c@{\hspace{8pt}}c@{\hspace{8pt}} c@{\hspace{8pt}}c@{\hspace{8pt}}c@{\hspace{8pt}}c@{\hspace{8pt}}c}
        \toprule
        Request Type & Data Type & GPT-3.5 & GPT-4 & GPT-4o & Gemini & Claude & Mixtral & Llama & Gemma \\
        \midrule
            \multirow{4}{*}{\shortstack{\bf Retrieval \\ (F1 score, \%)}}
            & Natural & 20.3 & 85.0 & 79.4 & 75.9 & \bf 76.5 & 59.1 & 77.5 & \underline{72.7} \\
            & Order-fixed & 20.1 & 83.6 & 81.5 & 82.2 & 73.1 & \underline{61.5} & \bf 82.6 & 71.5 \\
            & Template & \underline{27.8} & \underline{91.5} & \underline{84.5} & \underline{84.6} & \underline{74.1} & 58.5 & 73.4 & 70.0 \\
            & Table & \bf 30.1 & \bf 92.8 & \bf 88.6 & \bf 87.2 & 74.0 & \bf 63.7 & \underline{81.8} & \bf 74.0 \\
        \midrule
            \multirow{4}{*}{\shortstack{\bf Exclusion \\ (F1 score, \%)}}
            & Natural & 29.1 & \bf 70.1 & 62.9 & 56.3 & \bf 39.0 & 17.8 & \underline{51.1} & 19.0 \\
            & Order-fixed & 29.3 & 67.5 & \bf 70.6 & \underline{68.4} & 37.3 & 12.4 & 47.3 & 18.0 \\
            & Template & \underline{31.0} & 59.6 & \underline{68.8} & \bf 71.0 & 35.3 & \underline{20.4} & \bf 52.8 & \underline{27.5} \\
            & Table & \bf 38.8 & \underline{69.2} & \bf 70.6 & 63.5 & \underline{38.4} & \textbf{35.9} & 45.9 & \bf 40.7 \\
        \midrule
            \multirow{4}{*}{\shortstack{\bf Revision \\ (F1 score, \%)}}
            & Natural & 2.1 & \underline{27.4} & 30.0 & 29.8 & 2.4 & 1.2 & 6.4 & 1.9 \\
            & Order-fixed & 3.4 & 21.0 & \underline{32.5} & \underline{40.3} & 4.6 & 2.0 & 7.2 & 3.2\\
            & Template & \underline{7.9} & 20.1 & 29.5 & 30.5 & \underline{7.2} & \underline{3.1} & \bf 12.4 & \underline{4.0}\\
            & Table & \bf 17.4 & \bf 46.1 & \bf 55.0 & \bf 81.7 & \bf 15.9 & \bf 4.3 & \underline{7.9} & \bf 10.4\\
        \midrule
            \multirow{4}{*}{\shortstack{\bf Superlative \\ (Accuracy, \%)}}
            & Natural & \bf 9.2 & 37.2 & \underline{27.3} & \bf 31.3 & 76.2 & \underline{23.0} & \bf 28.8 & 14.2 \\
            & Order-fixed & \underline{8.5} & 41.8 & 24.0 & 25.3 & 75.8 & 21.8 & 26.3 & \bf 15.5 \\
            & Template & \bf 9.2 & \underline{44.8} & 26.2 & \underline{30.3} & \bf 82.0 & 21.5 & \underline{26.8} & \underline{15.2} \\
            & Table & 7.7 & \bf 46.8 & \bf 30.2 & 25.8 & \underline{80.2} & \bf 25.0 & 22.7 & 11.7 \\
        \midrule
            \multirow{4}{*}{\shortstack{\bf Summation \\ (Accuracy, \%)}}
            & Natural & 1.7 & 18.0 & 13.2 & 5.7 & 48.2 & 15.3 & 6.5 & 2.8 \\
            & Order-fixed & \underline{2.3} & 20.3 & \underline{13.0} & \underline{6.5} & 50.3 & \underline{16.2} & \underline{8.2} & \underline{3.5} \\
            & Template & 1.8 & \underline{22.3} & \bf 16.7 & 6.3 & \bf 54.7 & 14.2 & \bf 9.7 & \bf 6.0 \\
            & Table & \bf 3.7 & \bf 22.7 & \underline{13.0} & \bf 7.2 & \underline{51.0} & \bf 16.3 & 8.0 & 3.2 \\

        \midrule
            \multirow{4}{*}{\shortstack{\bf Quantification \\ (Abs. Diff., $\downarrow$)}}
            & Natural & 12.70 & 5.92 & 4.25 & \underline{5.80} & 2.24 & 12.50 & 10.74 & 10.25 \\
            & Order-fixed & 12.58 & 4.62 & 4.36 & 6.17 & 2.38 & 10.13 & 10.17 & 8.74 \\
            & Template & \bf 11.80 & \underline{4.28} & \bf 3.74 & 6.34 & \bf 1.17 & \underline{10.07} & \underline{9.95} & \underline{7.99} \\
            & Table & \underline{12.52} & \bf 3.20 & \underline{3.99} & \bf 4.44 & \underline{1.50} & \bf 7.98 & \bf 9.62 & \bf 7.00 \\
        
        \bottomrule
        \end{tabular}
    }
\end{table*}

\begin{table}[th]
    \caption{Extensive results of Fig.~\ref{fig:condition_portion} for partially structured information. The numbers are F1 scores, scaled into a percentage scale.}
    \label{tbl:portion}
    \centering
    \scalebox{1.0}{
    \begin{tabular}{c|c@{\hspace{10pt}}c@{\hspace{10pt}}c@{\hspace{10pt}}c}
    \toprule
     \multirow{2}{*}{Model}  & \multicolumn{4}{c}{Structuring Portion}\\
       & None & Quarter  & Half  & All \\
    \midrule
     GPT-3.5 & 20.3 & 18.8&20.1  & \bf 30.1\\ 
     GPT-4 & 85.0 & 88.5& 87.2 &\bf 92.8 \\ 
     GPT-4o & 79.4 & 85.0& 86.6 & \bf 88.6\\ 
     Gemini &  73.9 &87.9 & \bf 88.3 & 87.2\\ 
     Claude &  69.0 & 66.2 & 67.7 & \bf 69.4\\   
     Mixtral & 59.1  & \bf 64.2 & 62.5  & 63.7 \\ 
     Llama &  77.5 &  78.1  & 78.0 & \bf 81.8 \\ 
     Gemma &  72.7 & \bf 75.3& 75.2 & 74.0\\    
     \midrule
     Avg. & 67.1 & 70.5 & 70.7 & \bf 73.5\\
 
    \bottomrule
    \end{tabular}
    }
\end{table}

\begin{table*}[htb]
    \caption{Rationale accuracy for requests based on the existence operator on the \textit{Soccer} dataset. Note that the accuracy numbers are in decimal format.}
    \label{tbl:erbench}
    \centering
    \scalebox{0.85}{
    \begin{tabular}{c|c@{\hspace{8pt}}c@{\hspace{8pt}}c@{\hspace{8pt}} c@{\hspace{8pt}}c@{\hspace{8pt}}c@{\hspace{8pt}} c@{\hspace{8pt}}c}
    \toprule
     Data Type & GPT-3.5 & GPT-4 & GPT-4o & Gemini & Claude & Mixtral & Llama & Gemma \\
    \midrule
    \midrule
      Text & 0.36 & 0.85& 0.98& \textbf{0.97}&0.98 &0.92 &\bf0.98&0.88   \\ 
      Table & \textbf{0.70}& \textbf{0.94}& \textbf{0.99}&0.95 & \textbf{1.00}&\bf0.94&0.97& \bf 0.95  \\
     \cmidrule{1-9}
      \textbf{Text Negated} & 0.53 \textcolor{red}{(0.18)}& 0.89 \textcolor{red}{(0.04)}&0.98 \textcolor{red}{(0.00)}&0.90 \textcolor{red}{(0.07)}&0.97 \textbf{\textcolor{red}{(0.01)}} &  0.61 \textcolor{red}{(0.31)}&   \textbf{0.92} \textbf{\textcolor{red}{(0.06)}} &0.76 \textcolor{red}{(0.12)}  \\
     \textbf{Table Negated} &\textbf{0.75} \textbf{\textcolor{red}{(0.05)}} & \textbf{0.95} \textbf{\textcolor{red}{(0.01)}}&\textbf{0.99} \textcolor{red}{(0.00)}&0.90 \textbf{\textcolor{red}{(0.05)}}& 0.97 \textcolor{red}{(0.03)}&\textbf{0.90} \textbf{\textcolor{red}{(0.02)}}& 0.82 \textcolor{red}{(0.15)}& \textbf{0.83}  \textcolor{red}{(0.12)}  \\
    \bottomrule
    \end{tabular}
    }
  \end{table*}

\begin{table*}[!ht]
    \caption{Performance comparison requests with the set difference conditions on \textit{Soccer} dataset.}
    \label{tbl:minusoperator}
    \centering
    \scalebox{0.9}{
        \begin{tabular}{c|c|c@{\hspace{8pt}}c@{\hspace{8pt}}c@{\hspace{8pt}} c@{\hspace{8pt}}c@{\hspace{8pt}}c@{\hspace{8pt}}c@{\hspace{8pt}}c}
        \toprule
        Request Type & Data Type & GPT-3.5 & GPT-4 & GPT-4o & Gemini & Claude & Mixtral & Llama & Gemma \\
        \midrule
            \multirow{2}{*}{\shortstack{\bf Retrieval \\ (F1 score, \%)}}
            & Text & 46.5 & 85.9 & 83.4 & \bf 89.8 & \bf 82.7 & 66.6 & 72.4 & 70.2 \\
            & Table & \bf 58.1 & \bf 90.0 & \bf 89.7 & 85.6 & 76.3 & \bf 67.9 & \bf 81.8 & \bf 78.9 \\
      \midrule
            \multirow{2}{*}{\shortstack{\bf Summation \\ (Accuracy, \%)}}
            & Text & 3.7 & 15.7 & 12.0 & 2.7 & 50.0 & 11.7 & 7.3 & 1.7 \\
            & Table & \bf 9.3 & \bf 30.3 & \bf 17.0 & \bf 4.7 & \bf 61.3 & \bf 15.7 & \bf 11.0 & \bf 3.3 \\
        \midrule
            \multirow{2}{*}{\shortstack{\bf Quantification \\ (Abs. Diff., $\downarrow$)}}
            & Text & 5.48 & 3.70 & 3.51 & \bf 4.08 & 1.42 & 9.41 & \bf 3.90 & \bf 7.50 \\
            & Table & \bf 4.09 & \bf 3.11 & \bf 3.18 & 5.73 & \bf 1.31 & \bf 6.04 & 4.55 & 9.50 \\
        \bottomrule
        \end{tabular}
    }
\end{table*}

\begin{table*}[t]
\caption{Case study comparing LLM responses between cases where text and table are injected to the model, highlighting that structured data leads to more accurate and efficient results.}
    \label{tbl:casestudy_multiturn}
    \centering
    \small
    \begin{tabular}{c|p{0.8\linewidth}}
    \toprule
    Data Type & \textbf{CASE STUDY \scriptsize{FOR} \small Multi-turn Gemini} \\
    \midrule
    \midrule
        \multirow{7}{*}{Text}
        & \textbf{Q:} Count the number of soccer players, club name as Atlético de Madrid and league name as English Premier League. Answer in an integer form. \\\\
        & \textbf{A:} There are 2 players listed who play for Atlético de Madrid.  There are many players listed who play in the English Premier League.  I cannot provide a precise count without manually reviewing each entry again, which is inefficient.  A more efficient method would require a database. \\
    \midrule
        \multirow{5}{*}{\shortstack{Table}}
        & \textbf{Q:} Count the number of soccer players, club name as Atlético de Madrid and league name as English Premier League. Answer in an integer form. \\\\
        & \textbf{A:} There are 0 soccer players with club name Atlético de Madrid and league name English Premier League in the provided list. \\
    \bottomrule
    \end{tabular}
\end{table*}

\begin{table*}[th]
    \caption{Mapping requests to database query languages.}
    \label{tbl:promptrelationalalgebra}
    \centering
    \scalebox{0.85}{
    \begin{tabular}{c|c|c}
    \toprule
    Type&Language&Context\\
     \midrule
     \multirow{3}{*}{Retrieval} & \multirow{2}{*}{\bf Request}&Give me soccer players of nationality name as Germany\\ 
     &&and preferred foot as Right. \\
     \cmidrule(lr){2-3}
      &\textbf{Relational Algebra} & $\sigma_{\text{nationality} = \text{Germany} \land \text{preferred\_foot} = \text{Right}}(\text{\small Soccer})$\\
      \midrule
      \multirow{3}{*}{Exclusion} & \multirow{2}{*}{\bf Request}& Forget soccer players of nationality name as Germany \\&&and preferred foot as Right.\\
      \cmidrule(lr){2-3}
      &\textbf{Relational Algebra} & $\text{\small Soccer} := \text{\small Soccer} - \sigma_{\text{Nationality} = \text{Argentina} \wedge \text{Preferred\_Foot} = \text{Left}} (\text{\small Soccer})$\\
    \midrule
    \multirow{7}{*}{Revision}& &Update the names of soccer players to N/A, while leaving all  \\
    &\textbf{Request}&other attributes unchanged, if their nationality name is Germany  \\
    &&and their preferred foot is Right.\\
    \cmidrule{2-3}
    &\multirow{4}{*}{\textbf{Relational Algebra + SQL}} &$S1 \gets \sigma_{\text{nationality} = \text{Germany} \land \text{preferred\_foot} = \text{Right}}(\text{\small Soccer})$\\
    && $S2 \gets \sigma_{\text{nationality} \neq \text{Germany} \lor \text{preferred\_foot} \neq \text{Right}}(\text{\small Soccer})$\\
     &&  $S1 \gets$ $\text{\small UPDATE S1 SET name = ``N/A''}$\\
     && $Soccer \gets S1 \cup S2$ \\
    \midrule
     \multirow{3}{*}{Quantification} & \multirow{2}{*}{\bf Request}& Count the number of soccer players, nationality name as \\
     &&Germany and preferred foot not as Right.\\
     \cmidrule(lr){2-3}
     &\textbf{Relational Algebra}&$\text{\small COUNT}\left(\sigma_{\text{nationality} = \text{Germany} \land \text{preferred\_foot} = \text{Right}}(\text{\small Soccer})\right)$\\
     \midrule
     \multirow{3}{*}{Summation} & \multirow{2}{*}{\bf Request}& Sum the club jersey number of soccer players nationality \\&&name as Germany and preferred foot as Right.
     \\ \cmidrule(lr){2-3}
     &\textbf{Relational Algebra}&$\text{\small SUM}(\pi_{\text{jersey\_number}}\left(\sigma_{\text{nationality} = \text{Germany} \land \text{preferred\_foot} = \text{Right}}(\text{\small Soccer}))\right)$\\
     \midrule
     \multirow{9}{*}{Superlative} & \multirow{5}{*}{\bf Request}& Among soccer players nationality name as Germany and \\
     &&preferred foot as Right, give me one soccer player with the \\
     && highest uniform jersey number.If multiple players satisfy the \\
     &&condition, give me the player with the long name that comes \\
     &&first in alphabetical order.\\ 
     \cmidrule(lr){2-3}
     &\multirow{4}{*}{\bf Relational Algebra}&$S \gets \sigma_{\text{nationality} = \text{Germany} \land \text{preferred\_foot} = \text{Right}}(\text{\small Soccer})$\\
     &&$M \gets \text{\small MAX}\left(\pi_{\text{jersey\_number}}(S)\right)$\\
     &&$T \gets \sigma_{\text{jersey\_number} = M}(S)$\\
     &&$R \gets \sigma_{\text{name} = \text{\small MIN}\left(\pi_{\text{name}}(T)\right)}(T)$\\
     \midrule
     \multirow{4}{*}{Existence} & \multirow{2}{*}{\bf Request}&Is there a soccer player from Argentina who played for \\
     &&FC Barcelona with uniform number 10 in FC Barcelona?\\ \cmidrule(lr){2-3}
     &\multirow{2}{*}{\bf Relational Calculus}& $\exists t \in \text{\small Soccer} \, (t.\text{\small nationality} = \text{\small Argentina} \land t.\text{\small club} = \text{\small FC Barcelona}$ \\
     &&$\land t.\text{\small club\_jersey\_number} = \text{\small 10})$\\
     \midrule
     \multirow{3}{*}{Projection} & \multirow{2}{*}{\bf Request}&Provide me with soccer players' name and club jersey number of \\
     &&nationality name as Germany and preferred foot as Right. \\
     \cmidrule(lr){2-3}
      &\textbf{Relational Algebra} & $ \pi_{\text{name}, \text{club\_jersey\_number}} \left( \sigma_{\text{nationality} = \text{Germany} \wedge \text{preferred\_foot} = \text{Right}}(\text{\small Soccer}) \right)$\\

    \bottomrule
    \end{tabular}
    }
  \end{table*}

\end{document}